\documentclass[11pt, a4paper, logo, internal, copyright]{deepmind}

\usepackage[authoryear, sort&compress, round]{natbib}
\usepackage{booktabs}       
\usepackage{multirow}
\usepackage[group-separator={,}]{siunitx}

\title{Vision-Language Models as Success Detectors} 

\correspondingauthor{yuqing\_du@berkeley.edu, kksenia@deepmind.com, cabi@deepmind.com}

\keywords{success detection, vision language models, generalisation, reward models}

\setboolean{internal}{false}

\newcommand{\ours}{{\bf{FT Flamingo 3B}}}
\newcommand{\IA}{{\bf{IA Playroom}}}
\newcommand{\baseline}{{\bf{bespoke SD}}}

\author[2*]{Yuqing Du}
\author[1]{Ksenia Konyushkova}
\author[1]{Misha Denil}
\author[1]{Akhil Raju}
\author[1]{Jessica Landon}
\author[1]{Felix Hill}
\author[1]{Nando de Freitas}
\author[1]{Serkan Cabi}

\affil[1]{DeepMind}
\affil[2]{UC Berkeley}
\affil[*]{Work done during internship at DeepMind}

\begin{abstract}
Detecting successful behaviour is crucial for training intelligent agents. As such, generalisable reward models are a prerequisite for agents that can learn to generalise their behaviour. In this work we focus on developing robust success detectors that leverage large, pretrained vision-language models (Flamingo, \cite{alayrac2022flamingo}) and human reward annotations. Concretely, we treat success detection as a visual question answering (VQA) problem, denoted \textit{SuccessVQA}. We study success detection across three vastly different domains: (i) interactive language-conditioned agents in a simulated household, (ii) real world robotic manipulation, and (iii) ``in-the-wild" human egocentric videos. We investigate the generalisation properties of a Flamingo-based success detection model across unseen language and visual changes in the first two domains, and find that the proposed method is able to outperform bespoke reward models in out-of-distribution test scenarios with either variation. In the last domain of ``in-the-wild" human videos, we show that success detection on unseen real videos presents an even more challenging generalisation task warranting future work. We hope our initial results encourage further work in real world success detection and reward modelling.
\end{abstract}

\begin{document}
\maketitle


\section{Introduction}
\label{sec:intro}

Being able to detect successful (\textit{i.e.,} preferred) behaviour is a crucial prerequisite for training intelligent agents. For example, a signal of successful behaviour is necessary as a reward for policy learning, or as an evaluation metric for identifying performant policies. As such, in this work we are  concerned with developing accurate and generalisable \textit{success detectors}, which \textit{classify if a behaviour is successful or not}. While it is possible to engineer success detectors in specific domains, such as games \citep{mnih2013playing} or control tasks \citep{tunyasuvunakool2020}, in most real-world tasks they can be challenging to define. Success detection in realistic settings can be difficult not only due to challenges with identifying the environment state ({\it e.g.,} detecting a particular object configuration from pixels), but also due to ambiguities about what a successful state is ({\it e.g.,} subjective goals, such as ``generate an entertaining story''). 
One possible approach for developing success detectors is through reward modelling with preference data \citep{christiano2017deep, ouyang2022training, cabi2020sketchy, abbeel2004apprenticeship, ng2000algorithms}. However, the trained preference models are often accurate only for the fixed set of tasks and narrow environment conditions observed in the preference-annotated training data, and thus they require extensive labour-intensive annotations for better coverage. This presents a significant bottleneck, as we would like success detectors to be able to generalise broadly -- for instance, once a success detector learns what ``successfully picking up a block'' looks like, it should be able to detect this behaviour even if the background or agent morphology changes thanks to a semantic understanding of ``picking up a block''.

Consider success detection in robotic manipulation, where tasks are specified with language instructions and observations consist of images. We posit that generalisable success detection is useful for learning generalisable policies in this domain. Here, effective success detectors should generalise to task variations along two axes. Firstly, they should generalise to {\it language variations} in the task specification. For instance, a model that is trained on detecting success for the instruction ``lift a rubber duck'' should also accurately measure success for ``lift a toy duck object''. Secondly, success detectors should generalise to {\it visual variations}. For example, if a camera moves or additional objects are introduced in the scene, the model should still reliably detect success on accomplishing a known task. Standard reward models are typically trained for fixed conditions and tasks, and are thus unable to generalise to such variations. As such, adapting success detectors to new conditions typically requires collecting a new annotated dataset and re-training the model.

\begin{figure}
    \centering
    \includegraphics[width=1\textwidth]{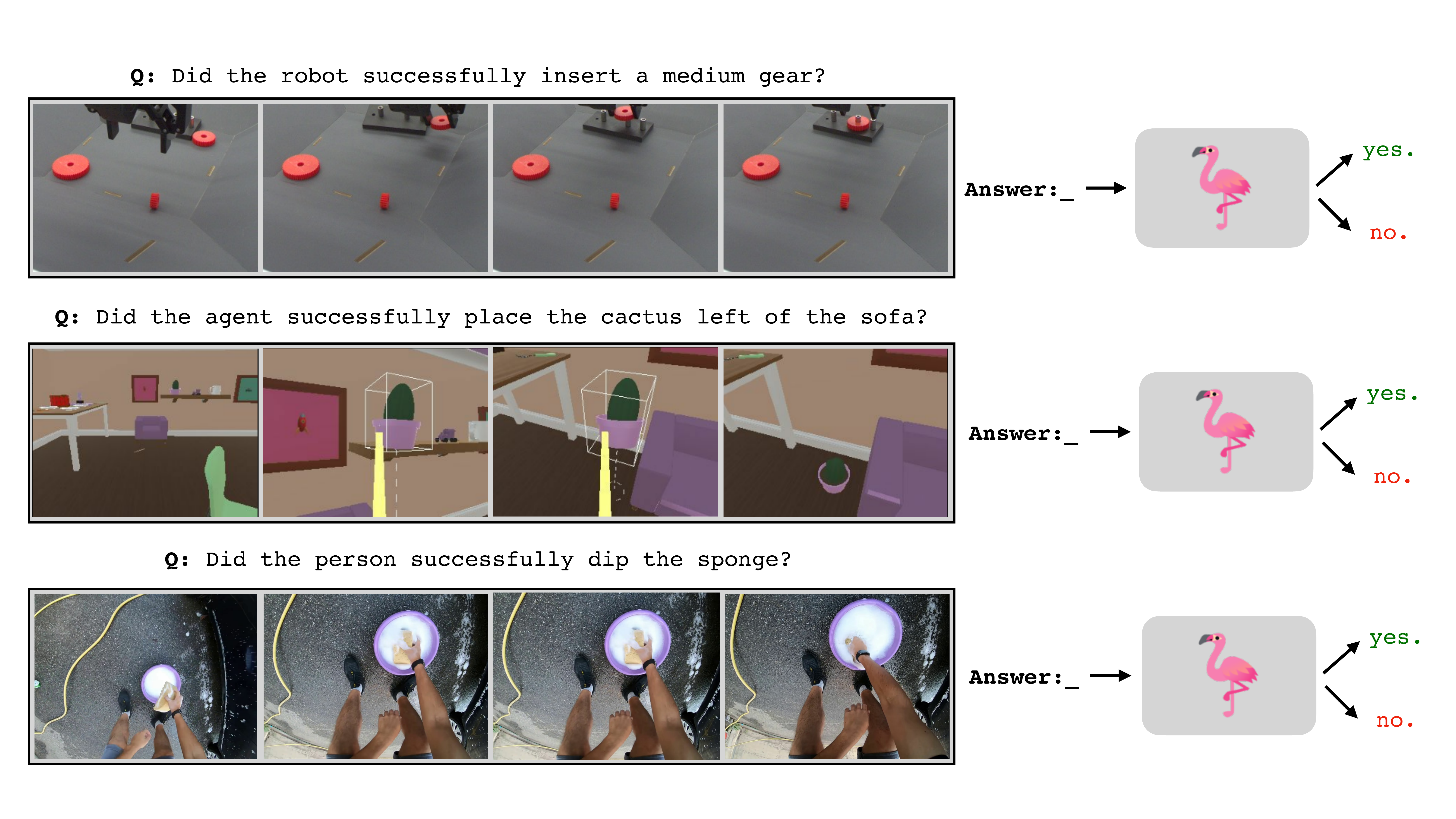}
    \caption{\textbf{SuccessVQA}: Success detection tasks can be formulated as visual question answering (VQA) problems. Large multimodal language models, such as Flamingo, offer the opportunity to learn a generalisable success detector, which can act either as a reward model or agent evaluator in a broad range of domains.} 
    \label{fig:successvqa} 
\end{figure}

In this work, we aim to train success detectors that are robust with respect to variations in both language specifications and perceptual conditions. To this end, we leverage large pretrained vision-language models (VLMs), such as Flamingo \citep{alayrac2022flamingo}, as a foundation for learning success detectors. We hypothesize that Flamingo's pretraining on vast amounts of diverse language and visual data will enable learning more robust success detectors. In particular, we show that the same simple approach of finetuning Flamingo with human annotations leads to generalisable success detection across vastly different domains.
This simple approach allows us to use a unified architecture and training scheme, where we require only 1) videos describing the world state, and 2) text describing the desired behaviour or task. We reframe the problem of success detection as a visual question answering (VQA) task and refer to this formulation as \textit{SuccessVQA} (Figure \ref{fig:successvqa}). 

Concretely, we finetune Flamingo for success detection on three diverse domains: a simulated household \citep{team2021creating}, real-world robotic manipulation, and in-the-wild egocentric human videos \citep{grauman2022ego4d}. The universality of the SuccessVQA task formulation is instrumental in enabling use of the same training architecture in a wide range of tasks and environments. We demonstrate that the resulting success detectors are capable of zero-shot generalisation to unseen conditions (both in language and vision) where bespoke learned reward models fail. 

\section{Related Work}

\paragraph{Vision-Language Models (VLMs)}  Multimodal vision-language models (VLMs) have shown remarkable success in recent years, where VLMs can serve as a foundation for various tasks using language, vision, or arbitrary combinations of modalities. VLMs can be trained with contrastive objectives \citep{jia2021scaling, radford2021learning} and/or generative objectives \citep{dai22enabling, hu21scaling, luo20uniVL, alayrac2022flamingo}. In this work we rely on the Flamingo model \citep{alayrac2022flamingo}, which leverages a contrastive objective for pretraining the vision encoder on text-and-image pairs. This is combined with a frozen pretrained language model though the Perceiver Resampler and interleaved cross-attention layers, and optimized with a generative objective. We approach success detection as a closed-form visual question answering (VQA) task. However, unlike other applications of VLMs in single-image VQA tasks \citep{tiong2022plug}, we rely on videos to specify the world state, making our work more similar to video QA tasks \citep{xu2016msr}. While the original Flamingo work demonstrates capabilities on video understanding, we extend this approach to training video-based reward models. Variants of our approach (\textit{e.g.,} by reducing the video input to a single frame) can also be applied with other VLMs built on large language models \citep{li2023blip, koh2023grounding}.

\paragraph{Reward Modelling} Reward modelling is often necessary when it is challenging to hard-code a reward function for an agent to learn from. To circumvent this, there has been a rich body of prior work on learning reward functions from data. When rewards are accessible through a simulator, one can use supervised learning to train a reward model for model-based agent learning \citep{hafner2023mastering}. However, many tasks can be difficult to simulate and hand-engineer simulated rewards for.  To overcome this challenge, one can learn reward models from human data. When demonstrations of desirable behaviour are available, one can leverage inverse reinforcement learning (IRL), where the key idea is to recover a reward function that best explains expert behaviour~\citep{ng2000algorithms,finn2016guided, ho2016generative, li2017infogail,fu2017learning, merel2017learning,zhu2018reinforcement,baram2017end}. However, IRL relies on access to such expert demonstrations, makes assumptions about the relationship between the expert actions and the true reward, and can be difficult to learn.

When demonstrations are difficult to acquire, a more natural way of providing human feedback is through comparative preferences that indicate the degree to which certain agent behaviour is desirable. This can be done with comparisons of whole episodes~\citep{akrour2012april, schoenauer2014programming, brown2019extrapolating, dorsa2017active}, trajectory segments~\citep{christiano2017deep,ibarz2018reward, lee2021pebble, abramson2022improving}, or even synthesized hypothetical trajectories~\citep{reddy2020learning}. These methods then fit a reward function as a preference-predictor, \textit{e.g.,} using a Bradley-Terry model \citep{bradley1952rank}. Nevertheless, preferences are not always the most natural form of feedback from humans, and in many cases we would like the exploit the goal-oriented nature of many tasks we care about. In other words, sometimes it can be easier for a person to provide direct success labels or scalar rewards with respect to a given goal. This can be done online in response to observed agent actions and state transitions \citep{knox2008tamer, macglashan2017interactive, arumugam2019deep}. In robotics, proposed methods vary from sparse, single frame annotations \citep{singh2019endtoend} to dense, whole trajectory annotations \citep{cabi2020sketchy}. In this work we learn from reward annotations, focusing on training success detectors which can be viewed as binary reward functions. Since collecting human annotations for each new task and environment can be expensive, we aim to study whether pretrained, large VLMs can enable learning more generalisable success detectors from human annotations.

\textbf{Large-Scale Pretraining for Success Detectors}
Our work falls under the general category of using foundation models as reward models. In language modelling, reward models are typically trained by finetuning a pretrained LLM with human preferences over LLM generations. This reward model can then be used to finetune an LLM with filtered supervised finetuning or reinforcement learning from human feedback (RLHF) \citep{stiennon2020summarize, nakano22webgpt, menick2022quotes, glaese22dialog, askell2021general, bai2022training}. For embodied agents, large-scale datasets of in-the-wild human videos have been used to train reward models \citep{ma2022vip, chen2021learning}. Rather than using human reward annotations of agent behaviours, these methods rely on task-annotated human videos of successful behaviours.  Most similar to our work, some prior approaches propose using contrastive VLMs as reward models. In simulated robot domains, \cite{mahmoudieh2022zero, cui2022can} propose using CLIP \citep{radford2021learning} to generate task rewards from a text-based goal description and pixel observations. \cite{fan2022minedojo} leverage large-scale Minecraft data to finetune a Minecraft-specific video CLIP model for detecting alignment ({\it i.e.,} reward) with text task descriptions. Our work differs in that we leverage a generative VLM built on a frozen large language model, which we hypothesize enables better language generalisation. We also apply our method to three vastly different domains, including real-world domains where ground truth rewards are difficult to obtain, and thus directly make use of human annotations.

\section{SuccessVQA: Success Detection as a VQA Task}

Our primary contribution is SuccessVQA, a framework that allows us to train multi-task success detectors by directly leveraging powerful pretrained VLMs, such as Flamingo. In SuccessVQA, the VLM is given a visual input representing the state of the world ({\it e.g.,} a single image or a short video clip) and a question asking if the specified task is successfully accomplished. This problem formulation has several advantages:
\begin{itemize}
    \item It allows us to unify success detection across domains, using the same architecture and training scheme. We consider three domains: a simulated 3D playroom used in prior research on language-conditioned interactive agents  (\IA{})~\citep{team2021creating, abramson2020imitating}, real robotic manipulation, and ``in-the-wild" human videos from Ego4D~ \citep{grauman2022ego4d}.
    \item Relying on a pretrained vision-language model enables us to harness the advantages of pretraining on a large multimodal dataset. We hypothesize that this is the reason for better generalisation to both language and visual variations.
    \item The task and state specification allows us to unify treatment of success detection in tasks defined either by singular successful states or target behaviours ({\it i.e.,} detecting success requires reasoning across multiple frames).
\end{itemize}

\paragraph{SuccessVQA Datasets}
\begin{figure}[t]
    \centering
    \includegraphics[width=\textwidth]{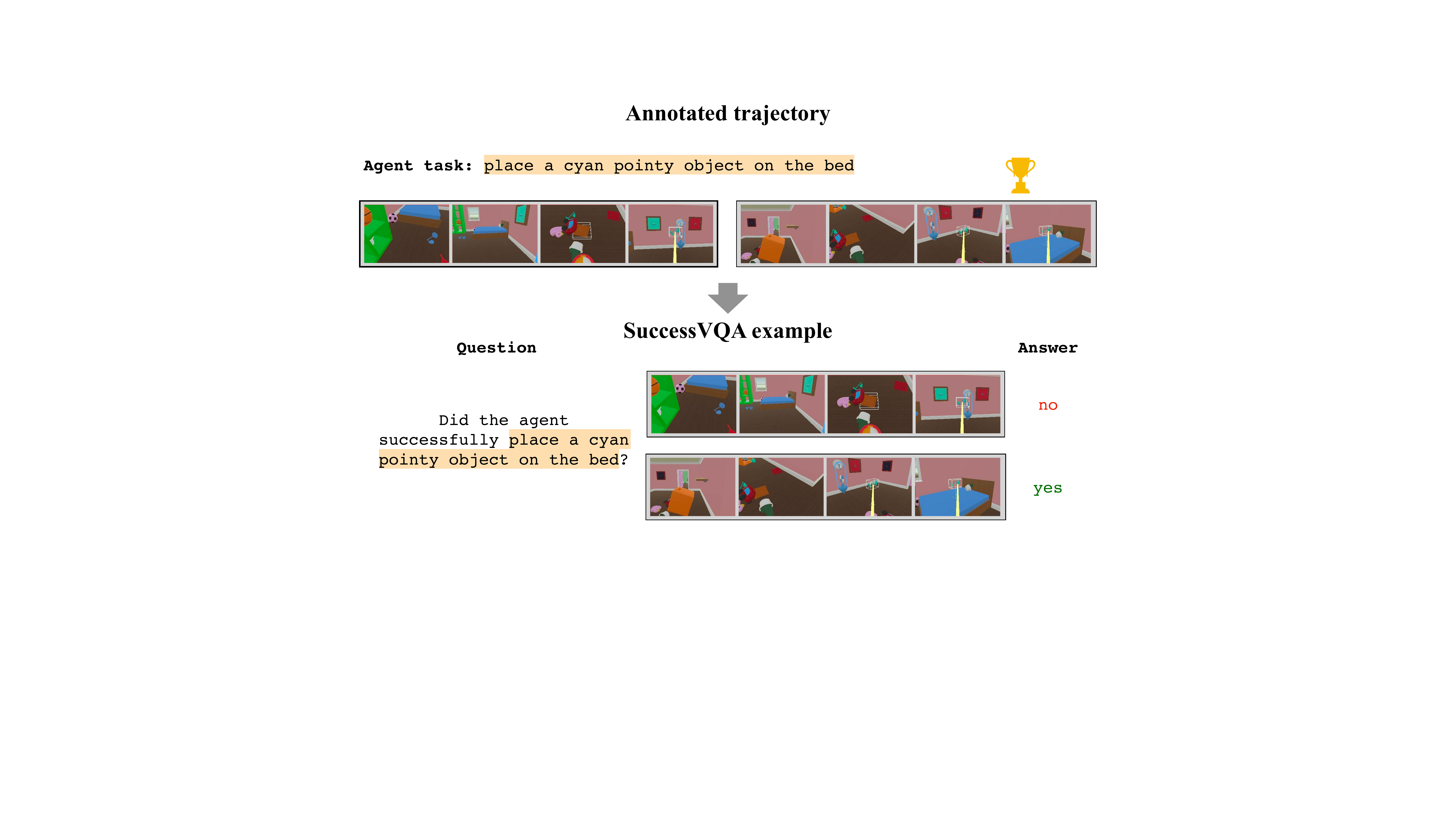}
    \caption{{\bf SuccessVQA dataset creation:} A trajectory is annotated by human raters with a point of success (denoted by the trophy). Then the trajectory is split into subsequences and converted to multiple SuccessVQA datapoints with corresponding questions and answers.}
    \label{fig:iasuccessvqa}
\end{figure}
To create the SuccessVQA datasets, we use behaviour trajectories annotated by humans to indicate  \textit{whether} a task is completed successfully, and if so, \textit{when} a success occurs. There may be multiple annotations per trajectory from different human raters. In the cases where raters disagree, success or failure is determined by a majority vote, and the median (across the raters who annotated success) of the first annotated success frame is used as the 'point of success'. All subsequent frames are also successful, unless the task is reversed (\textit{e.g.} removing a gear after inserting it for the robotics domain). To generate SuccessVQA examples, a trajectory is split into non-overlapping subsequences (Figure \ref{fig:iasuccessvqa}). For simplicity, we make the clip lengths the same as the pretraining clip lengths used for Flamingo: by first creating subsequences of length 211 frames, then downsampling from $30$ FPS to $1$ FPS to create $8$-frame subsequences. We then generate the VQA question using one of two methods. When trajectories correspond to some known task, we use the template: \texttt{``Did the robot/agent/person successfully \{task\}?''}, for example, \texttt{``Did the agent successfully place the cactus left of the sofa?''} (see Figure \ref{fig:successvqa}, first and second rows). When no task is provided but there is a narration corresponding the  actions in the clip, as in Ego4D, we use a frozen Flamingo model to rephrase the narrations into questions. For example, given a narration \texttt{``The person is scooping the ice cream''}, we convert it to the question \texttt{``Did the person successfully scoop the ice cream?''} (see Figure \ref{fig:successvqa}, last row). Finally, the answer is generated: \texttt{``yes''} if the given subsequence ends in success frames, and \texttt{``no''} otherwise. 

\paragraph{Training and Evaluation}

\begin{figure}[t!]
    \centering
    \includegraphics[width=\textwidth]{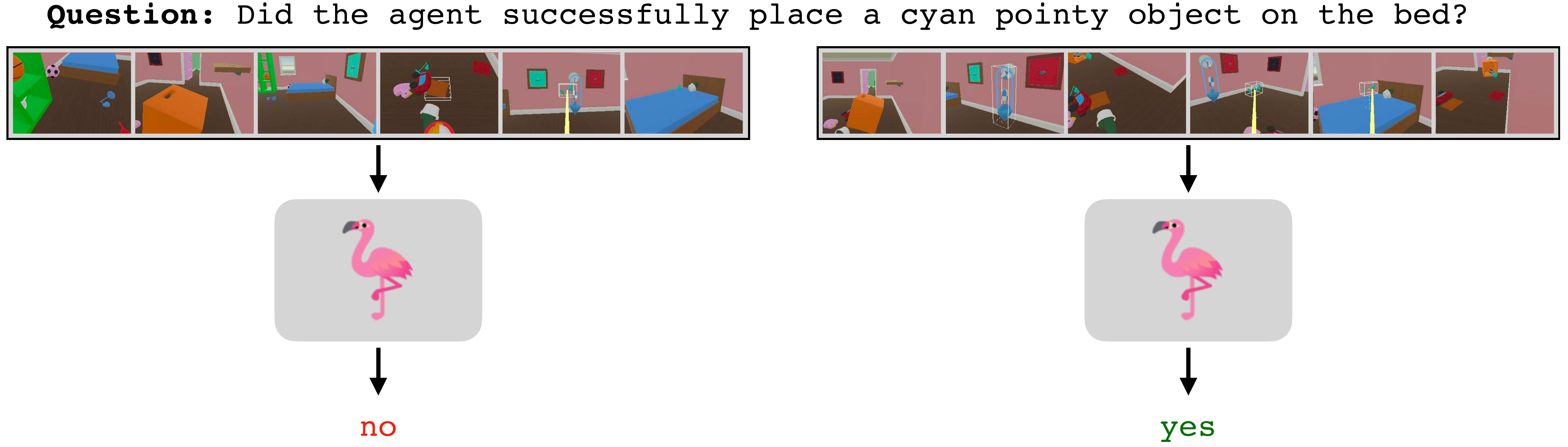}
    \caption{We compute episode-level success detection accuracy during evaluation in order to compare against bespoke success detection models for each domain. To do this, we create subsequences and predict success on each clip individually, then consolidate the predictions at an episode level.}
    \label{fig:eval_procedure}
\end{figure}
 
We finetune the Flamingo (3B) vision-language model on the SuccessVQA dataset for each domain. Specifically, we finetune all the vision layers (vision encoder, perceiver, and cross attention layers) and keep the language layers frozen. In the experiments we refer to this model as the \ours{}. For evaluation we compute clip-level success detection accuracy against the ground truth human annotations on held-out trajectories. In the simulated household and robotics domains (Sections \ref{sec:IA} and \ref{sec:robotics}) we also compute episode-level accuracy to directly compare against baseline bespoke success detection models, denoted \baseline{}. Note that these baselines were hand-designed independently and tuned specifically for each domain. While these models differ from Flamingo in both pretraining schemes and architecture, they represent \textit{a best attempt at designing an accurate reward model for in-distribution evaluations}. Episode-level success detection is computed as follows: first, we generate subsequences from the test trajectories in the same way as during training. Next, the success detection model classifies each clip individually for success, as illustrated in Figure \ref{fig:eval_procedure}. We consolidate the classifications in one of two ways. 1) When the success is completely defined by the observed environment state (as in the robotics tasks), we only look at the first and the last clip of an episode. Then, the entire episode as successful if the first clip is in a failure state and the last clip is in a success state. 2) When the success is defined by a particular behaviour (as in the simulated household domain), if any subsequence in an episode is classified as success we classify the episode as successful. We report {\emph{balanced}} accuracy on the test episodes, as there can be a large imbalance between the number of successful and failure episodes in the dataset. A random model would achieve 50\% balanced accuracy.

\paragraph{Experiments overview}

We use the SuccessVQA problem formulation to train success detectors across a diverse range of tasks in vastly different domains: simulated household or \IA{} (Section \ref{sec:IA}), robotics (Section \ref{sec:robotics}), and Ego4D videos (Section \ref{sec:Ego4D}). We investigate whether Flamingo as a success detector model backbone enables generalisation across the following axes:
\begin{itemize}
    \item {\bf language generalisation} (Section \ref{sec:IA}). Can we accurately detect success for novel tasks specified with language? To answer this question, we evaluate generalisation to unseen tasks specified with language. For example, if we train on detecting success for the task \texttt{``arrange objects in a row''}, can we accurately detect success for the task \texttt{``arrange objects in a circle''}? For these experiments, we use simulated tasks in the \IA{} environment where the trajectory dataset contains a large and diverse set of language-specified tasks.
    \item {\bf visual robustness} (Section \ref{sec:robotics}). Can we detect success in the presence of unseen visual variations? To answer this question, we evaluate success detection accuracy for a known semantic task, but in the presence of naturalistic visual perturbations. In these experiments, we use real-world robotic manipulation tasks where we introduce visual variations at test-time using different camera viewpoints and distractor objects.
\end{itemize}
We compare our model against bespoke evaluation models designed and trained specifically for each domain. We do not necessarily expect the Flamingo-based models to outperform the bespoke models in a given in-distribution scenario. Rather, we aim to investigate whether the Flamingo-based models have better robustness to both aforementioned language and visual changes, while also not requiring any domain-specific architectural or training changes. We emphasize that the benefit of SuccessVQA is the simple task formulation that can be applied across a wide range of domains and is directly amenable for use with large pretrained VLMs. Finally, in Section \ref{sec:Ego4D} we show an example of an in-the-wild SuccessVQA dataset derived from Ego4D \citep{grauman2022ego4d}. Initial results for success detection in this domain are promising, and we hope to encourage further work on accurate reward modelling in unstructured real-world settings.
\section{Language Robustness with Interactive Agents (\IA{})} \label{sec:IA}

In this section we train and evaluate success detectors in the simulated \IA{} environment, a diverse 3D house environment designed for training language-conditioned interactive agents \citep{team2021creating, abramson2020imitating}. The environment consists of ``a randomised set of rooms, with children’s toys and domestic objects, as well as containers, shelves, furniture, windows, and doors" (see Figure 1 in \cite{abramson2020imitating}). The tasks are generated from human-human interactions in the \IA{}, where a \textit{setter} is instructed to provide a task via language for a \textit{solver}, \textit{e.g.,} \texttt{"bring me the book from the living room"}. Success detectors in this environment can serve as automated evaluators for trained policies.

There are two properties in this environment that are particularly challenging for automated success detection: large language variety and the environment's multi-task nature. Large language variations are present because the tasks were originally generated from human interactions, and people are likely to use diverse language to specify even semantically similar tasks. For example, the task of bringing an object to the setter can be phrased in many ways: \texttt{"bring a fruit from the pantry"}, \texttt{"bring me the banana which is in the pantry"}, \texttt{"bring the yellow coloured object near me"}. Moreover, success detection in this environment is intrinsically \emph{multi-task} in its nature because: (1) there is a vast set of possible tasks that can be specified with different utterances, and (2) the behaviour of different people and trained agents can vary greatly for the same task. For automated evaluation, it is not scalable to train a new model for each language and task variation. 

\subsection{Methodology}

\begin{figure}
    \centering
    \includegraphics[width=0.6\textwidth]{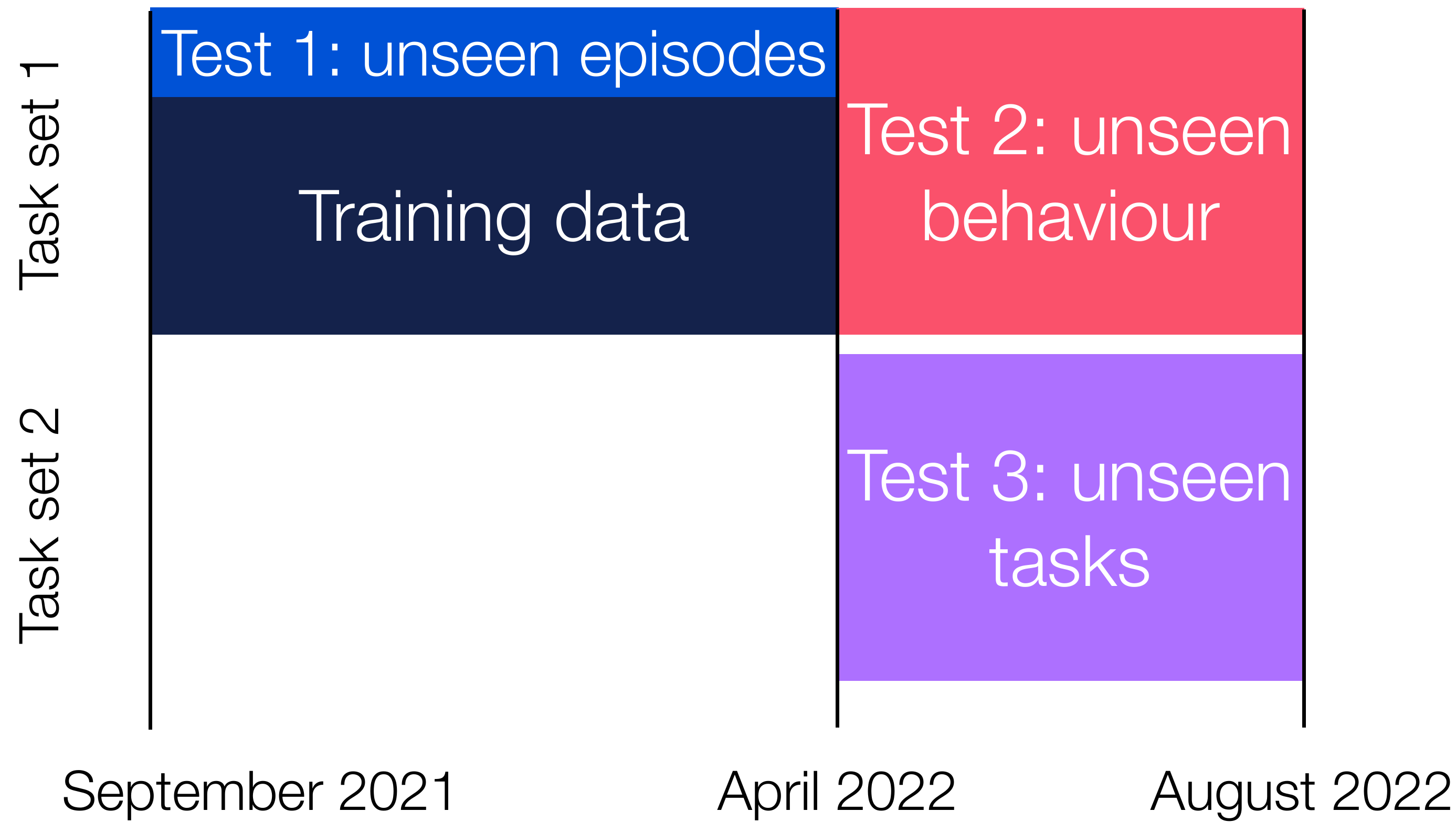}
    \caption{{\bf \IA{} datasets split.} Training and evaluation splits for \IA{} STS evaluation. Test episodes include unseen trajectories, unseen behaviours, and/or unseen language instructions. See Table \ref{tab:ia_taskset2} for details on the different task sets.}
    \label{fig:ia_setup}
\end{figure}

\paragraph{Training Dataset}
We use tasks and trajectories from the Standardized Test Suite (STS), designed specifically for evaluating learned Interactive Agents \citep{team2021creating, abramson2020imitating}. We focus on the movement-based tasks: tasks that require the solver agent to move around and interact with the environment. The STS consists of a set of "scenarios that typify the behaviour [the Interactive Agents team] wishes to evaluate" \citep{abramson2022evaluating}, and various trained agent policies are tasked with accomplishing the given scenarios. These test episodes are then annotated by human raters to indicate if a task is successfully completed and if so, at which frame success occurred. We use these annotations to create a SuccessVQA dataset for \ours{} finetuning and to train a \baseline{} model for comparison. The training set consists of STS and human interaction data collected between September 2021 to April 2022 (see Figure \ref{fig:ia_setup}), 546,887 trajectories in total (1,421,111 clips).

\paragraph{Baseline Success Detectors}
For the \baseline{} baseline, we use a success detection model specifically designed for the STS independently of this work.  There are two types of baseline models: whole episode evaluation and autoregressive evaluation. As the whole episode model consistently outperformed the autoregressive model, in this section we only report the results from that baseline (see Appendix \ref{app:sts} for additional results). This model creates a downsampled set of \num{32} frames from the entire evaluation episode and embeds the images with a ResNet-101. The agent input and output text are embedded using a learned text embedding. All embeddings are then concatenated together and fed to a transformer with an MLP head that predicts the likelihood the episode was successful. In addition to training on success detection, an auxiliary instruction-matching contrastive loss is applied. 

\paragraph{Evaluation} 
To select the best success detection model, we use the model and checkpoint with the highest balanced accuracy on a held-out validation split from the same distribution as the training data. We then evaluate the chosen success detector model across three different test sets: 
\begin{itemize}
    \item \textbf{Test 1: unseen episodes (in distribution)} -- a randomly held-out 10\% of training dataset trajectories, which includes rephrasings of training tasks. This dataset contains 175,952 clips. 
    \item \textbf{Test 2: unseen behaviour (out of distribution agents)} -- trajectories generated by {\emph{new agents}} on tasks seen in the training dataset, including rephrasings of training tasks. These agents potentially demonstrate novel behaviour. This allows us to assess success detector robustness to unseen behaviours on known tasks, which is important as it determines if we can reuse the same models even as agent behaviour evolves over time (\textit{i.e.} the success detector should be accurate even when the agent solves a known task in a novel way). 
    This dataset contains 462,061 clips. 
    \item \textbf{Test 3: unseen tasks (out of distribution tasks and agents)} -- the most challenging setting: trajectories generated by new agents on {\emph{new tasks}} not seen during training. For examples of how these tasks differ from the training set, see Table \ref{tab:ia_taskset2}. Note that this set comprises \textit{completely new tasks} as well as rephrasings of said tasks.  As the tasks are new, the success detector models need to master a semantic understanding of language to properly generalise to success detection in this set. This dataset contains \num{272031} clips.
\end{itemize}

\begin{table}[]
    \centering
    \footnotesize
    \begin{tabular}{c|p{12cm}}
    \toprule
        Property &  Test 3 Examples\\ \hline
        Unseen descriptor & \texttt{``arrange 4 pointy objects in a square shape in the bed room"}, where \texttt{square} is not mentioned in the training set. Instead, at train time we have tasks arranging objects in an \texttt{arc} or \texttt{triangle}. \\ \midrule
        Unseen objects & \texttt{``push the train engine with water bird"}, where neither \texttt{train engine} nor \texttt{bird} are mentioned in the training set. \\ \midrule
        Unseen actions & \texttt{``hit the candle using the pillow which is left of airplane in the living room"}, where the action \texttt{hit} is not mentioned in the training set. \\ \bottomrule
    \end{tabular}
    \caption{Examples of unseen task variants from Task Set 2, used in Test 3.}
    \label{tab:ia_taskset2}
\end{table}

\subsection{Experimental Results}

\begin{table}[ht!]
    \centering
    \footnotesize
    \begin{tabular}{c|c|c|c}
    \toprule
        Model & Test 1: & Test 2: & Test 3:  \\
         & unseen episodes & unseen behaviour & unseen tasks  \\ \midrule

        \baseline{} & 80.6\% & \textbf{85.4\%} & 49.9\% \\ 
        \ours{} & \textbf{83.4\% }& 85.0\% & \textbf{59.3\%} \\\bottomrule
    \end{tabular}
    \caption{Zero-shot episode-level balanced accuracies for \IA{} evaluation models. For reference, human level balanced accuracy is around \num{88}\% due to inter-rater disagreement.}
    \label{tab:ia_eval}
\end{table}

Table \ref{tab:ia_eval} presents the episode-level balanced accuracy on each test set. We find that without finetuning, the accuracy of the Flamingo model is close to random chance (see Appendix \ref{app:sts} for details). This is unsurprising, as the IA domain differs greatly from Flamingo's pretraining data. With finetuning on the same training set, \ours{} matches the performance of \baseline{} in both Test 1 (unseen episodes) and Test 2 (unseen behaviour). More importantly, in Test 3 (unseen tasks), the performance of a bespoke model drops to a random chance, while \ours{} outperforms it by a significant margin ($10$\%), see Table \ref{tab:ia_eval}. 
As the instructions in Test 3 are for novel tasks, not just rephrasings of tasks seen during training, this experiment demonstrates that the success detector exhibits some amount of semantic understanding of the scenes. We hypothesize that this is possible due to Flamingo's large language model backbone and web-scale pretraining. That said, there is still a large margin for improvement on the most challenging test set. For future work, it would be interesting to investigate how different model scales, dataset sizes, or cross-finetuning with different datasets can affect generalisation.
\section{Visual Robustness with Robotic Manipulation} \label{sec:robotics}

\begin{figure}[ht]
    \centering
    \includegraphics[scale=.2]{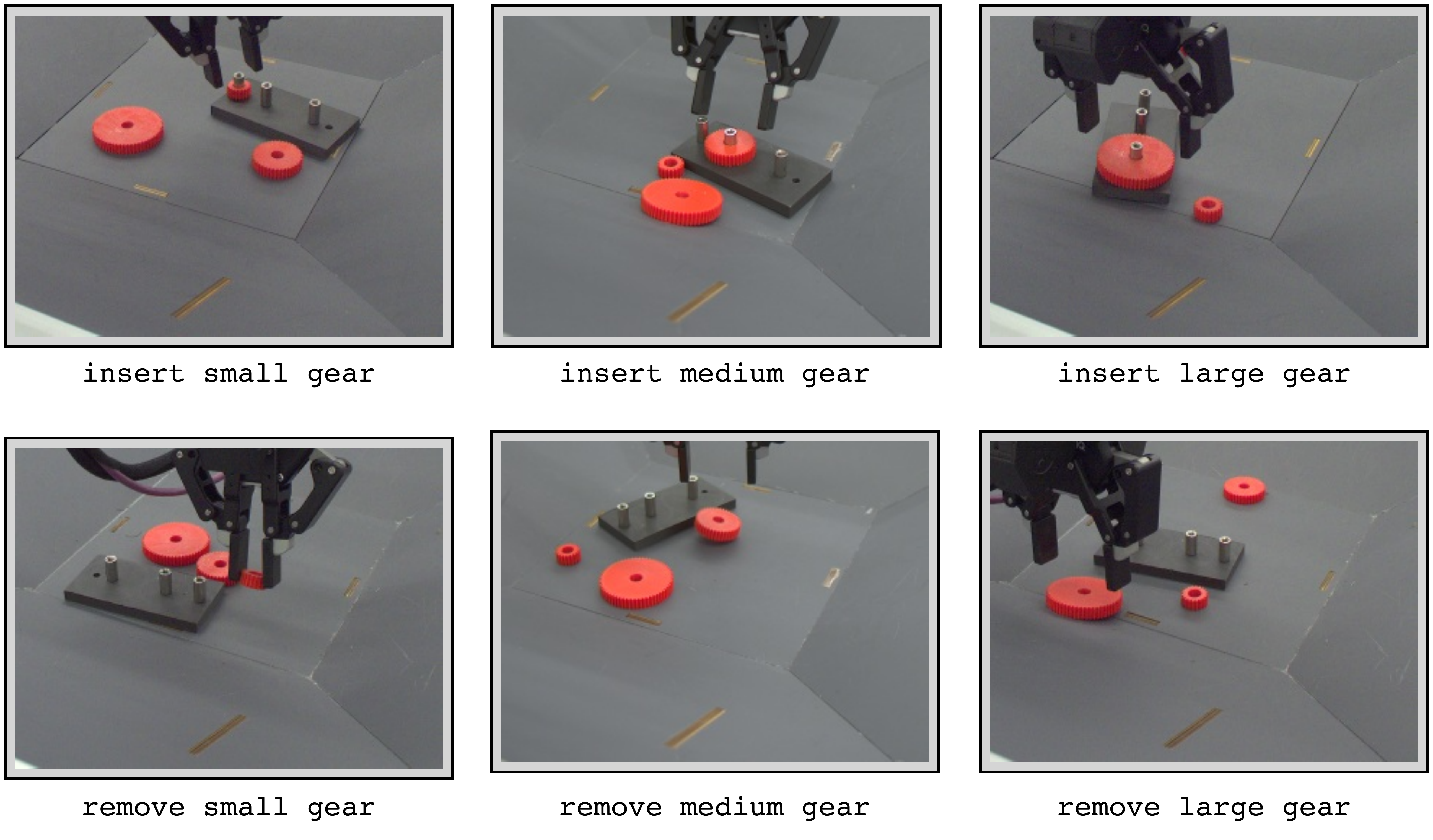}
    \caption{Successful frames for the $6$ robotics gear manipulation tasks: insert small, medium, large, and remove small, medium, large gear.}
    \label{fig:robotics_tasks}
\end{figure}

In this section we train and evaluate success detectors on a family of real-life robotic gear manipulation tasks with a Panda robot arm. There are six tasks corresponding to inserting or removing a small, medium, or large gear within a basket (Figure \ref{fig:robotics_tasks}). We consider visual observations from a basket camera. 
Ideally, a success detector should remain accurate under naturalistic visual changes, such as different camera view angles, lighting conditions, or backgrounds. Furthermore, as the performance of learned policies improves, we may want to introduce new objects or tasks to the environment. It quickly becomes impractical to re-annotate and re-train success detectors from previous tasks in new conditions, thus making it important to train visually robust success detectors. For example, a model that has learned to detect successful gear insertion should still be able to robustly detect success even if the basket has additional task-irrelevant distractor objects or the camera angle changes. To investigate this, we experiment with zero-shot evaluations on episodes with such visual changes.

\subsection{Methodology}
\begin{figure}
    \centering
    \includegraphics[width=\textwidth]{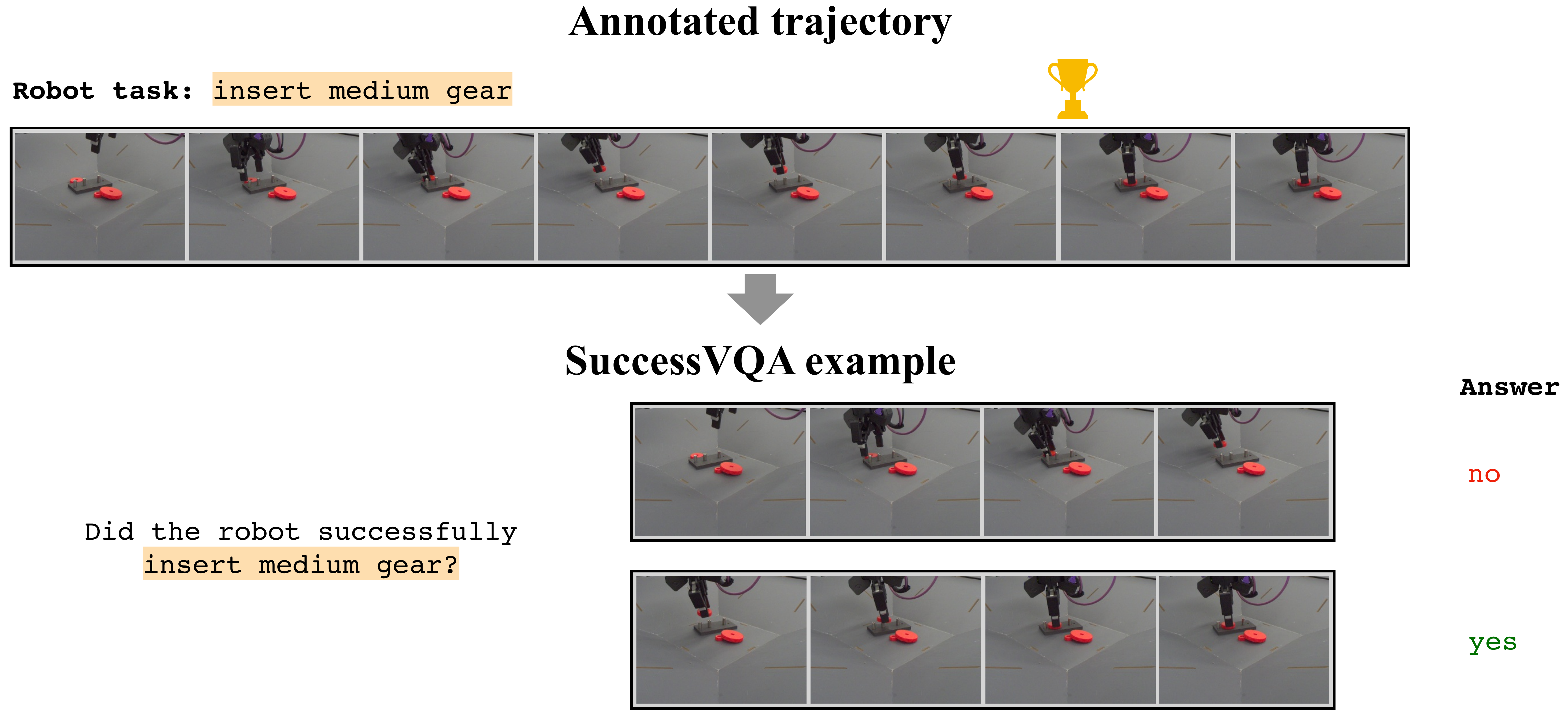}
    \caption{Sample SuccessVQA example created from an annotated subsequence of a gear manipulation episode. Success annotation is shown with the trophy.}
    \label{fig:reboticsVQA}
\end{figure}

\paragraph{Training dataset}
Human operators provide $101,789$ demonstrations for $6$ tasks using a 6DoF control device. Each episode is then annotated by humans with rewards for each task ({\it e.g.,} every episode has $6$ reward annotations, one for each task). Human annotators label positive rewards for all frames with a success state (\textit{i.e.,} if the task is solved), and zero rewards otherwise. Note that it is possible for a task to be accidentally undone in the same episode, at which point the reward annotation would revert to zero. The reward annotations and corresponding episode frames are then converted into SuccessVQA examples (see Figure \ref{fig:reboticsVQA}). The ground truth VQA answer is obtained from the human annotations: clip answers are labelled successful if they contain only a single transition from zero to positive reward or only have positive rewards throughout, otherwise they are labelled as unsuccessful. We train a single \ours{} success detector model for all \num{6} tasks.

\paragraph{Baseline Success Detector}

As a baseline, we consider a ResNet-based \citep{he2016deep} per-frame success classification model, tuned specifically for this task by the robotics team. The ResNet-18 is pretrained on ImageNet, and the classification layer is swapped out for a binary classification layer. We finetune a \textit{separate} success classification model for each of the \num{6} gear tasks, with image augmentations applied during training. This is distinct from our method where we train a single multi-task model across all \num{6} conditions. We consider an episode successful if the first and last frames\footnote{We find that incorporating more frames does not improve episode-level accuracy.} of the episode are classified as a failure (output < $0.5$) and success (output > $0.5$) correspondingly. We will further refer to the baseline model as bespoke success detector (\baseline{}). 

\paragraph{Evaluation}
To compare against the \baseline{}, we look at episode-level balanced accuracy. Given an evaluation episode, we consider the episode successful under \ours{} if the first clip is classified as unsuccessful and the last clip is classified as successful (see Figure \ref{fig:GTrobotics} in the Appendix). This matches the episode-level classification scheme of \baseline{}.

\begin{figure}
    \centering
    \includegraphics[width=\textwidth]{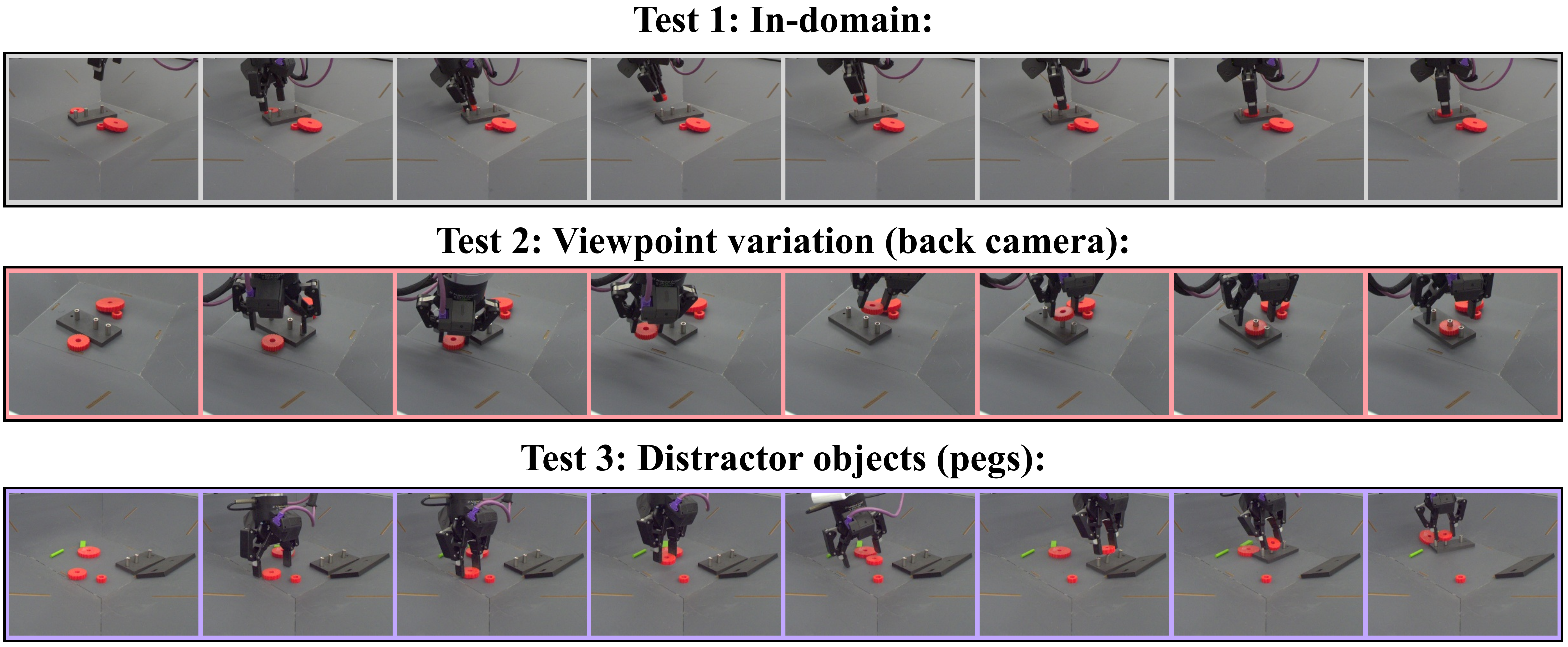}
    \caption{Examples of three evaluation datasets: in-domain episodes similar to the training dataset, episodes with a different camera viewing angle and episodes with distractor objects in the basket.}
    \label{fig:test_sets}
\end{figure}

We conduct the evaluation on three test sets (see Figure \ref{fig:test_sets}):
\begin{itemize}
    \item Test 1: In-domain episodes (first row),
    \item Test 2: Episodes with a {\bf viewpoint variation}, using a different (back) camera (second row),
    \item Test 3: Episodes with {\bf distractor objects} in the basket, but the original camera (last row).
\end{itemize}
The last two settings are designed to test the robustness of the models to naturalistic visual perturbations in the environment. The trained success detectors can then either be used as automated evaluators or reward models for agent training.

\subsection{Experimental results}

\subsubsection*{In-Domain Performance}\label{subsec:indomain}


In Test 1, we conduct an in-domain evaluation where the test set comes from the same visual conditions as the training set (see Figure \ref{fig:test_sets}, top row). The test set includes all the training episodes and an additional held out \num{2076} episodes. 
The results in Table \ref{tab:robotics_indomain} show that while the \baseline{} consistently outperforms the \ours{}, the performance of the \ours{} model is still comparable for the insertion task. Note that the accuracy of the Flamingo model on the remove tasks is lower, which we hypothesize is likely due to a data balancing issue. We have \num{5} times more training data available for insertion than removal, and training a single model across all tasks likely led to a tradeoff in accuracy between the insertion and removal tasks, which are temporal opposites of each other.


\begin{table}[ht]
    \centering
    \footnotesize
    \begin{tabular}{c|ccc|ccc}
    \toprule
         & \multicolumn{3}{c|}{Insert} & \multicolumn{3}{c}{Remove} \\ 
         & Small & Medium & Large & Small & Medium & Large \\ \midrule
        \baseline{} & \textbf{98.0\% }& \textbf{98.4\%} & \textbf{99.1\% } & \textbf{97.3\% }& \textbf{98.7\%} & \textbf{98.4\%} \\
        \ours{} & 96.0\% & 94.4\% & 95.0\% & 82.1\% & 83.4\% & 87.2\% \\ \bottomrule
    \end{tabular}
    \caption{{\bf In-Domain Episode-level Accuracy for Gear Manipulation.} Balanced accuracy evaluated on \num{50000}-\num{60000} episodes per task.}
    \label{tab:robotics_indomain}
\end{table}


We further verify that the {\bf FT Flamingo 3B} success detector can be used to train useful policies using reward-filtered behaviour cloning (BC). In filtered BC, we first use {\bf FT Flamingo 3B} to classify demonstration episodes as successes or failures for a particular task. Then, we use only the episodes classified as success for BC training. Table \ref{tab:robotics_policies} shows the average success rates of the policies evaluated on \num{20} episodes with manual resets. In manual resets no extra gears are pre-inserted on the pegs for the insert task and only the one relevant gear is pre-inserted for the remove tasks. The success rates vary between \num{50}\% and \num{75}\%, suggesting that the accuracy of the success detector models is sufficient for some amount of policy training. To compare with the \baseline{} model, we also conduct filtered BC training with the \baseline{} reward model and evaluate an insert large gear policy over \num{100} episodes with automated resets. In automated resets, policies for different tasks are run in sequence one after another and any number of gears might be already inserted at the start of the episode, presenting a harder scenario. In this case, the success rate is \num{30}\%  with \ours{} is and \num{33}\% with \baseline{}. This provides a preliminary proof-of-concept that the difference in reward model accuracy does not lead to a large difference in policy performance. We leave more detailed policy evaluations to future work.

\begin{table}[ht]
    \centering
    \footnotesize
    \begin{tabular}{c|ccc}
    \toprule
         & Small & Medium & Large \\ \midrule
        Insert & $55$\% & $65$\% & $70$\% \\
        Remove & $60$\% & $75$\% & $60$\% \\
        \bottomrule
    \end{tabular}
    \caption{{\bf Policy success rates.} Policies are trained with filtered behaviour cloning where only successful episodes are used for training and success is determined by {\bf FT Flamingo 3B}.}
    \label{tab:robotics_policies}
\end{table}

\subsubsection*{Visual Robustness}\label{subsec:visualrobustness}

Next, we focus on testing the generalisation capabilities of success detectors. We measure zero-shot accuracy on two natural visual variations described above: Test 2 and Test 3.

In Test 2, we look at zero-shot robustness to different viewpoints (Figure \ref{fig:test_sets}, middle row). Given that the success detectors were only trained on frames from the {\it front} basket camera, we evaluate robustness by measuring success detector accuracy on episodes recorded with the {\it back} basket camera. As we can see in Table \ref{tab:zero_shot_camera}, changing the camera angle drastically hurts the quality of \baseline{} (accuracy decreases of \num{10}-\num{50} absolute percentage points) while the performance of \ours{} is more stable (accuracy decreases by less than $10$\%). Note that in some tasks the performance of the bespoke model drops to the level of random guessing, essentially rendering the model useless for success detection. With this, {\bf FT Flamingo 3B} becomes the best performing model in $5$ out of $6$ tasks.

\begin{table}[ht]
    \centering
    \footnotesize
    \begin{tabular}{c|ccc|ccc}
    \toprule
         & \multicolumn{3}{c|}{Insert} & \multicolumn{3}{c}{Remove} \\ 
         & Small & Medium & Large & Small & Medium & Large \\ \midrule
        \multirow{2}{*}{\baseline{}} & 78.0\% & 53.1\% & 50.9\% & \textbf{85.8\%} & 53.8\% &  72.8\% \\
        & \textcolor{red}{{\bf -19.9\%}} & \textcolor{red}{{\bf-45.4\%}} &  \textcolor{red}{{\bf-48.3\%}} & \textcolor{red}{{\bf-11.5\%}} & \textcolor{red}{{\bf-44.9\%}} & \textcolor{red}{{\bf-25.5\%}} \\ \midrule
        \multirow{2}{*}{\ours{}} & \textbf{91.0\%} & \textbf{89.8\%} & \textbf{89.7\%} & 76.7\%  & \textbf{75.9\%} & \textbf{79.4\%} \\
         & \textcolor{red}{-4.0\%} & \textcolor{red}{-4.6\%} & \textcolor{red}{-5.3\%} & \textcolor{red}{-5.5\%} & \textcolor{red}{-7.5\%} & \textcolor{red}{-7.8\%} \\ \bottomrule
    \end{tabular}
    \caption{{\bf Viewpoint variation.} Zero-shot success detection balanced accuracy when trained on the front camera view and evaluated on the back camera view. We show the absolute balanced accuracy and the percentage point change compared to Test 1 from Table \ref{tab:robotics_indomain}.}
    \label{tab:zero_shot_camera}
\end{table}

\begin{table}[ht]
    \centering
    \footnotesize
    \begin{tabular}{c|ccc|ccc}
    \toprule
         & \multicolumn{3}{c|}{Insert} & \multicolumn{3}{c}{Remove} \\ 
         & Small & Medium & Large & Small & Medium & Large \\ \midrule
        \multirow{2}{*}{\baseline{}} & 88.8\% & 85.0\% & 71.8\% & \textbf{93.6\%} & \textbf{93.8\%} & \textbf{92.4\%} \\
        & {\bf \textcolor{red}{-9.2\%}} & {\bf \textcolor{red}{-13.4\%}} & {\bf \textcolor{red}{-27.4\%}} &  {\bf \textcolor{red}{-3.8\%}} & {\bf \textcolor{red}{-4.9\%}} & {\bf \textcolor{red}{-6.0\%}} \\ \midrule
        \multirow{2}{*}{\ours{}} & \textbf{96.1\%} & \textbf{95.6\%} & \textbf{90.6\%} & 82.4\% & 83.6\% & 84.7\% \\
         & \textcolor{green}{+0.1\%} & \textcolor{green}{+1.2\%} & \textcolor{red}{-4.5\%} & \textcolor{green}{+0.3\%} & \textcolor{green}{+0.1\%} & \textcolor{red}{-2.5\%} \\ \bottomrule
    \end{tabular}
    \caption{{\bf Distractor Objects.} Zero-shot success detection balanced accuracy on scenes with distractor objects. We show the absolute balanced accuracy and the percentage point change compared to Test 1 from Table \ref{tab:robotics_indomain}.}
    \label{tab:zero_shot_pegs}
\end{table}

Next, in Test 3 we look at zero-shot robustness in the setting where some \emph{distractor objects} (two pegs and a board, see Figure \ref{fig:test_sets}, last row) are introduced. Table \ref{tab:zero_shot_pegs} shows that detecting success on known tasks across this novel visual setting causes a $4$-$30$\% (absolute percentage points) drop in balanced accuracy for the bespoke model, while the accuracy mostly stays stable for the Flamingo-based models, with a $4.5$\% drop in accuracy at most. 

These two experiments demonstrate that Flamingo-based success detection models are robust to natural visual variations. We hypothesize that the pretrained Flamingo-based success detection model is better suited to zero-shot visual generalisation than the bespoke baseline reward model, as Flamingo is pretrained on a diverse set of visual data with corresponding language grounding. While the baseline model was also pretrained and used image augmentations during task finetuning, it was not exposed to such a diverse set of visual data or language. Large-scale diverse pretraining might contribute to better semantic tasks recognition under naturalistic visual changes. These encouraging results suggest that pretrained VLM-based success detectors are likely better suited to the real-world tasks involving unstructured, open, and evolving settings.

\section{Real World Success Detection with Ego4D} \label{sec:Ego4D}

In this section we describe creating a SuccessVQA dataset using ``in-the-wild" egocentric videos of humans performing tasks. This present a much more diverse setting than the prior two domains, in both visuals and language. We construct this dataset using annotations from the Ego4D dataset \citep{grauman2022ego4d}, where unlike prior benchmarks in action recognition, the focus is on detecting a temporal point of success for a given action. It is an example of a realistic, unstructured setting where the ground-truth success labels can be obtained only from human annotations. While the \ours{} success detector model shows initial promising results, our experiments show that the benchmark is nonetheless very challenging with much room for future progress.

\begin{figure}[ht!]
    \centering
    \includegraphics[width=.8\textwidth]{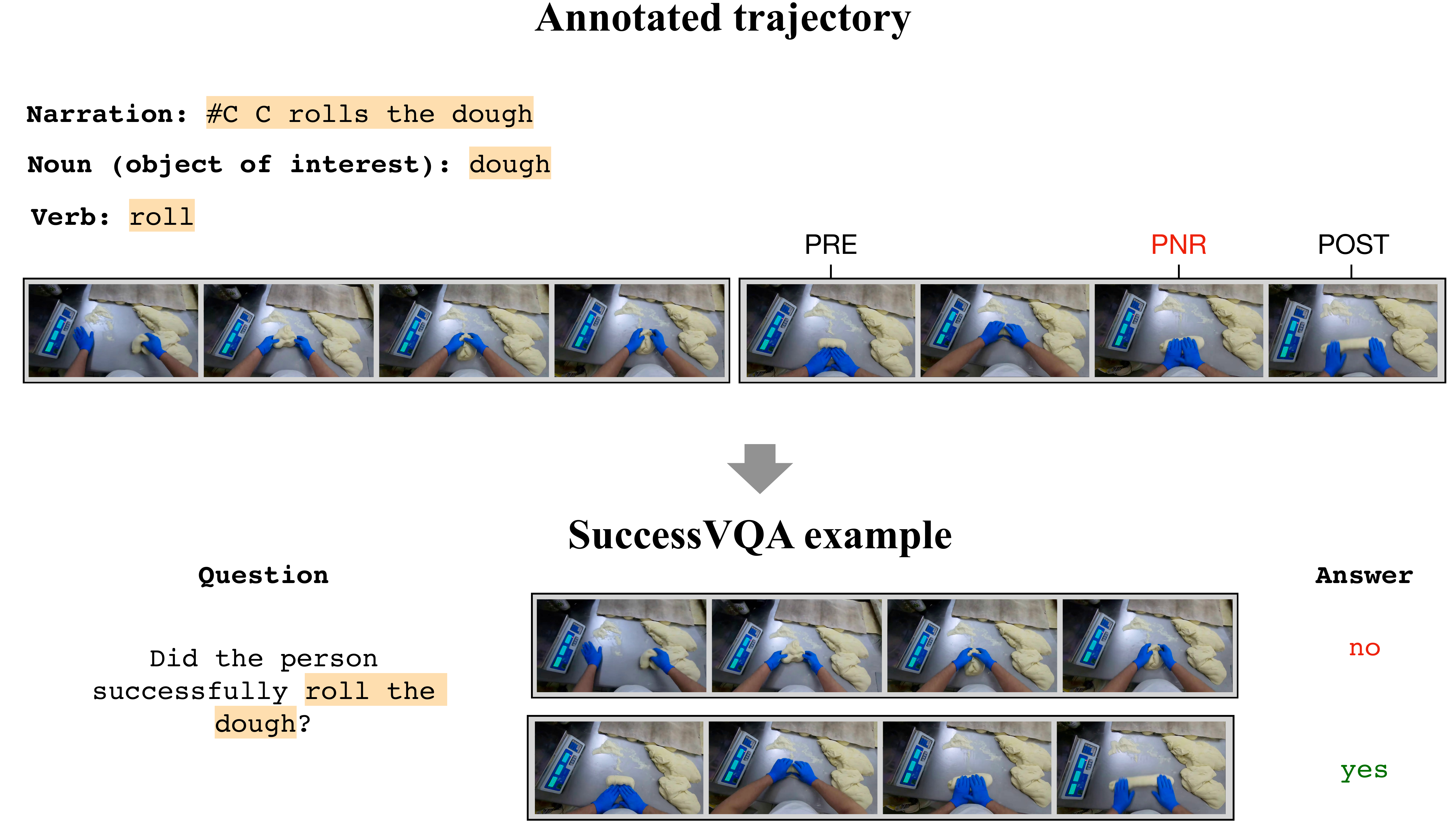}
    \caption{Sample Ego4D clip converted to SuccessVQA Examples. Ego4D annotations include PRE, POST and PNR (point of no return) annotations which are then used to generate answers in the SuccessVQA examples.}
    \label{fig:ego4d}
\end{figure}

Ego4D is a publicly available dataset of egocentric human-in-the-wild videos. The videos show people executing common tasks (\textit{e.g.,} washing dishes, cleaning cars, gardening). To generate ``successful" and ``unsuccessful" action sequences, we make use of annotations from the Ego4D Forecasting + Hands \& Objects (FHO) dataset, where corresponding narrations describe the actions of the camera wearer in the videos. Additionally, critical state changes are annotated: ``how the camera wearer changes the state of an object by using or manipulating it--which we call an object state change'' \citep{grauman2022ego4d}. Each narration is centered on an $8$-second clip, which is further annotated with action verbs, object nouns, and state change types corresponding to the narration and clip, as well as the critical frames PRE, Point of No Return (PNR), and POST for indicating when the state change has occurred. The PNR frame annotates the start of the state change, the PRE frame indicates a point before the state change, and the POST frame is a point after the state change is completed. 

We propose using the critical frame annotations as annotations of ``success" for the behaviour described in the narration. Specifically, we treat PNR frame as a point at which ``success" occurs. To generate a negative example for a clip, we use the frames in the $8$-second clip prior to the PRE frame. These frames do not contain the point of success, but they often demonstrate the beginning of the relevant action. We then generate the questions for SuccessVQA by rephrasing the narrations into questions using Flamingo, as shown in Figure \ref{fig:ego4d}. 

Unlike the \IA{} and robotics domains where there is only one relevant task per episode, a single Ego4D ``episode" (\textit{i.e.} video) can have multiple narrations corresponding to different actions. Thus, instead of episode-level accuracy we evaluate success detection accuracy on clips taken from \textit{held out videos}. In our experiments, \ours{} finetuned on the SuccessVQA dataset attains  99\% training balanced accuracy and  62\% test set balanced accuracy. For context, zero shot and 4-shot Flamingo models only achieve 50\% and 52\%.
That is, without finetuning, the Flamingo model is not capable of detecting success. Providing a few examples with few-shot prompting improves performance, but very slightly. However, finetuning Flamingo on the in-domain Ego4D SuccessVQA examples achieves a significant improvement over random chance. That said, there is still a large gap between train and test performance. We find that it is currently difficult to generalise to completely unseen videos and language tasks, so this domain provides an exciting avenue for future work.

\section{Conclusion}

In this work we propose SuccessVQA -- a reformulation of success detection that is amenable to pretrained VLMs such as Flamingo. We investigate success detection across a wide range of domains: simulated language-conditioned interactive agents, real-world robotic manipulation, and ``in-the-wild" human videos. We find that the pretrained VLM has comparable performance on most in-distribution tasks, and increased robustness across language and visual changes compared to task-specific reward models, and emphasize that our contribution is a more universal success detection task formulation that can be applied easily across vastly different domains.  VLMs can be used as policies, see {\it e.g.,} \cite{reed2022generalist}, but in this work we have demonstrated that there is also great value in using them as reward models. In contrast to VLMs as policies, VLMs as rewards focuses on the `what to do' and not on `how to do it'. We therefore expect such models to transfer more easily than policies when the same task can be accomplished in many ways, and where fine visual details are not necessary (\textit{e.g.,} grasp angle for fine motor control).

That said, this method has some limitations. There still exist some gaps between the Flamingo-based reward models and the bespoke reward models in our experiments, especially in some tasks in the robotics environment. Furthermore, inference with a larger VLM is expensive, making online success detection challenging. Lastly, we find that finetuning on a sufficient amount of in-domain data is necessary for robust success detection, as zero-shot or few-shot performance is not sufficient yet. Nonetheless, we are optimistic that further progress on broadly improving VLMs will result in more accurate few-shot success detection.

To address the limitations of the current approach, improving inference speed or distillation to a smaller model can help with efficient online success detection. Before deployment as a reward model for learning policies, we need further investigations into model accuracy and thorough characterizations of the effects of false positives and false negatives. So far we have experimented with a Flamingo 3B, but larger models might bring further improvements in robustness and generalisation. Another interesting avenue would be to investigate the practicality of in-domain few-shot generalisation to novel tasks (\textit{e.g.,} train on `insert gear' tasks, then detect success on `remove gear' after prompting with a few examples). An interesting question is \textit{when} to choose few-shot or finetuning and how to combine the two. The shared SuccessVQA format can enable shared finetuning across different datasets (\textit{e.g.,} combining Ego4D SuccessVQA and VQAv2 \citep{goyal2017making}) to study the impact of cross-task transfer. Lastly, the flexibility in input format of VLM models allows us to consider success detection tasks where the task is specified visually ({\it e.g.,} with a goal image) or the state is described in language ({\it e.g.,} a dialogue agent) in the same framework as the current work.

\section*{Acknowledgements}

We would like to thank Olivia Watkins and Antoine Miech for careful proofreading of the paper and detailed comments. We would also like to thank the DM Robotics Team, the Interactive Agents team, and the Flamingo team for insightful discussions and research support.

\bibliography{string,main}
\pagebreak

\appendix

\section{Simulated household domain}\label{app:sts}

To evaluate agent policies on the standardized set of scenarios (STS), each agent is first given a period of context to replay up to a "continuation point", after which the agent policy is used to complete the trajectory. Each continuation is then evaluated offline by human annotators as either successful or failure, along with the point at which success or failure occurs. These human annotations are then used to rank agent policies, using the proportion of successful annotations they receive. For more details on the evaluation procedure, see \cite{abramson2022evaluating}.

\subsection{Baseline Evaluation Models}
While human evaluations provide the ground truth signal for assessing agent capabilities, the cost of annotations scales directly with the number of evaluations for each new task and agent. Thus, there has been interest in automating the evaluation protocol to enable evaluation to scale over time. Ideally, an automated evaluation model will condition on an episode of agent behaviour and the input task utterance, and output a classification whether or not the task is successful. 

Currently two baseline evaluation models have been developed for the STS: whole-episode and autoregressive models. In both cases, the reward annotations for a particular episode are aggregated using majority voting.  

\textbf{Whole episode evaluation models.}
\begin{figure}[ht!]
    \centering
    \includegraphics[width=\textwidth]{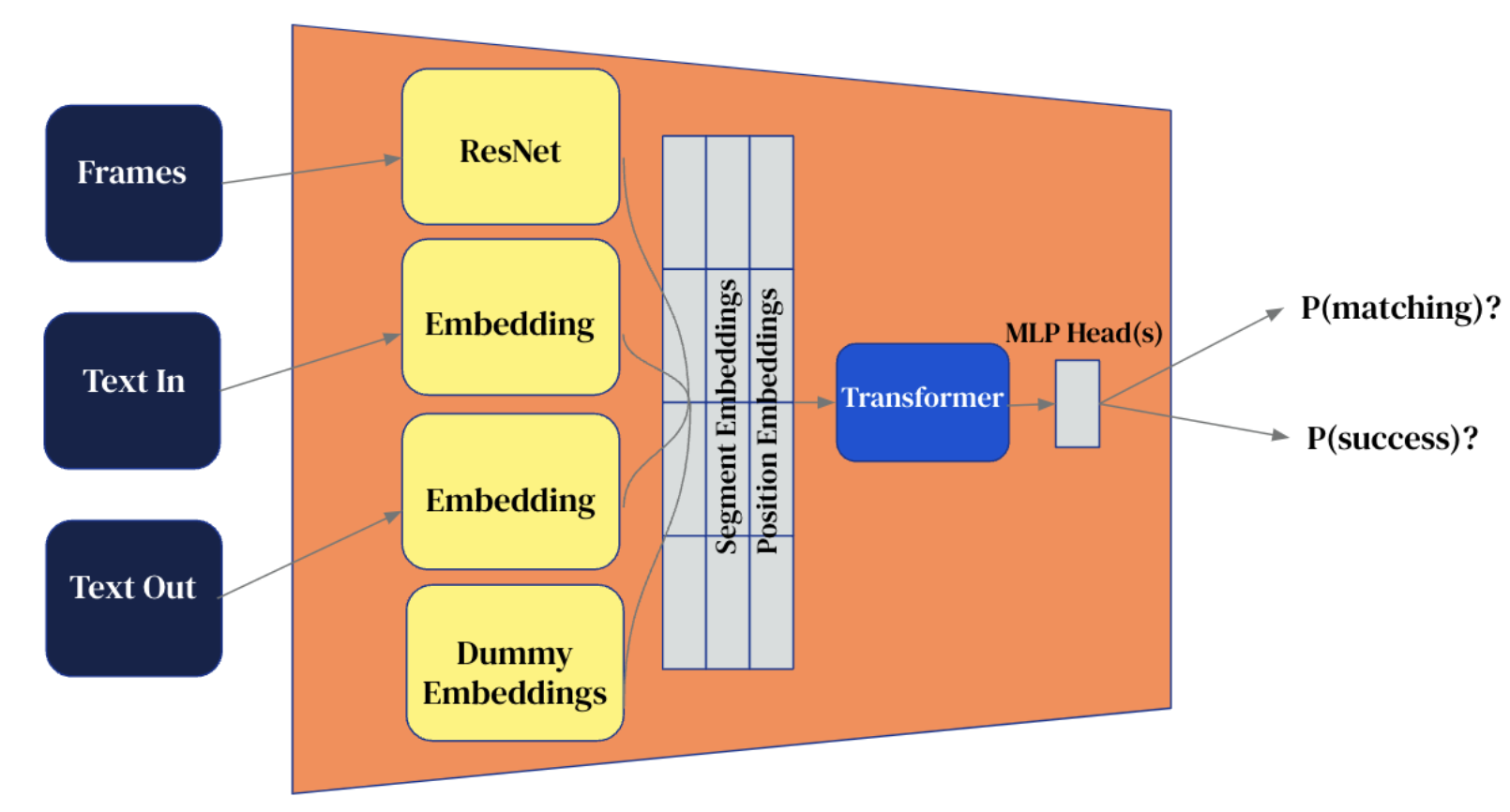}
    \caption{Whole Episode Bespoke Evaluation Model}
    \label{fig:IAWholeEp}
\end{figure}

For these models, we first preprocess an STS episode by downsampling it to 32 frames and tokenizing the text instruction and agent responses. The images are then embedded with a ResNet-101, the input and output text are embedded, and these embeddings are concatenated together and fed to a transformer with 16 layers and 16 attention heads. The transformer output is fed through two MLP heads: one to predict the likelihood of the episode being successful, P(success), and an auxiliary contrastive loss, P(matching). P(success) is supervised with the aggregated reward annotations, and P(matching) is trained to predict whether an instruction matches the episode or has been shuffled.  

\textbf{Autoregressive evaluation models.}
\begin{figure}[ht!]
    \centering
    \includegraphics[width=\textwidth]{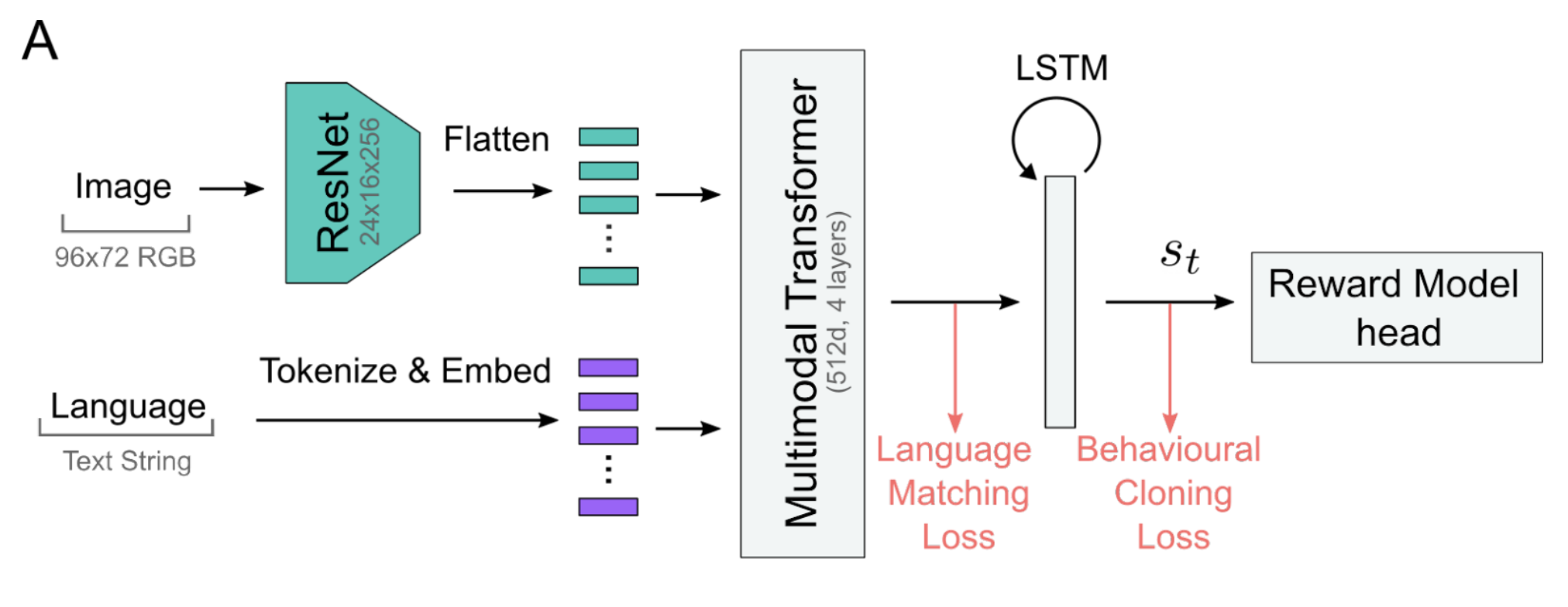}
    \caption{Autoregressive Bespoke Evaluation Model}
    \label{fig:IAWAR}
\end{figure}
The autoregressive evaluation models use the same architecture as the Playhouse agents, which takes inputs on a per-frame basis, rather than at the episode level. The model embeds the images and language for each frame, passes the embeddings to a multimodal transformer followed by an LSTM, and is asked to predict success or no-success on a per frame basis. The success of an entire episode is then determined by whether or not any single frame was predicted to be successful.

\begin{table}[h!]
    \centering
    \begin{tabular}{c|c|c|c}
        Model & Test 1: & Test 2: & Test 3:  \\
         & unseen episodes & unseen behaviour & unseen language  \\ \hline

        Baseline Whole Episode Model & 80.6\% & \textbf{85.4\%} & 49.9\% \\ 
        Baseline Autoregressive Model & 71.7\% & 70.4\% & (not tested) \\  \hline
        Flamingo 3B & 50\% & 50\%  & 50\%  \\
        \ours{} & \textbf{83.4\% }& 85.0\% & \textbf{59.3\%}
    \end{tabular}
    \caption{Zero-shot episode-level balanced accuracies for \IA{} STS evaluation models. For reference, human level balanced accuracy is around 88\%.}
    \label{tab:ia_eval_extra}
\end{table}
\newpage
\section{Robotics domain}
\subsection{Ground truth in robotics domain}

\begin{figure}[ht!]
    \centering
    \includegraphics[width=0.5\textwidth]{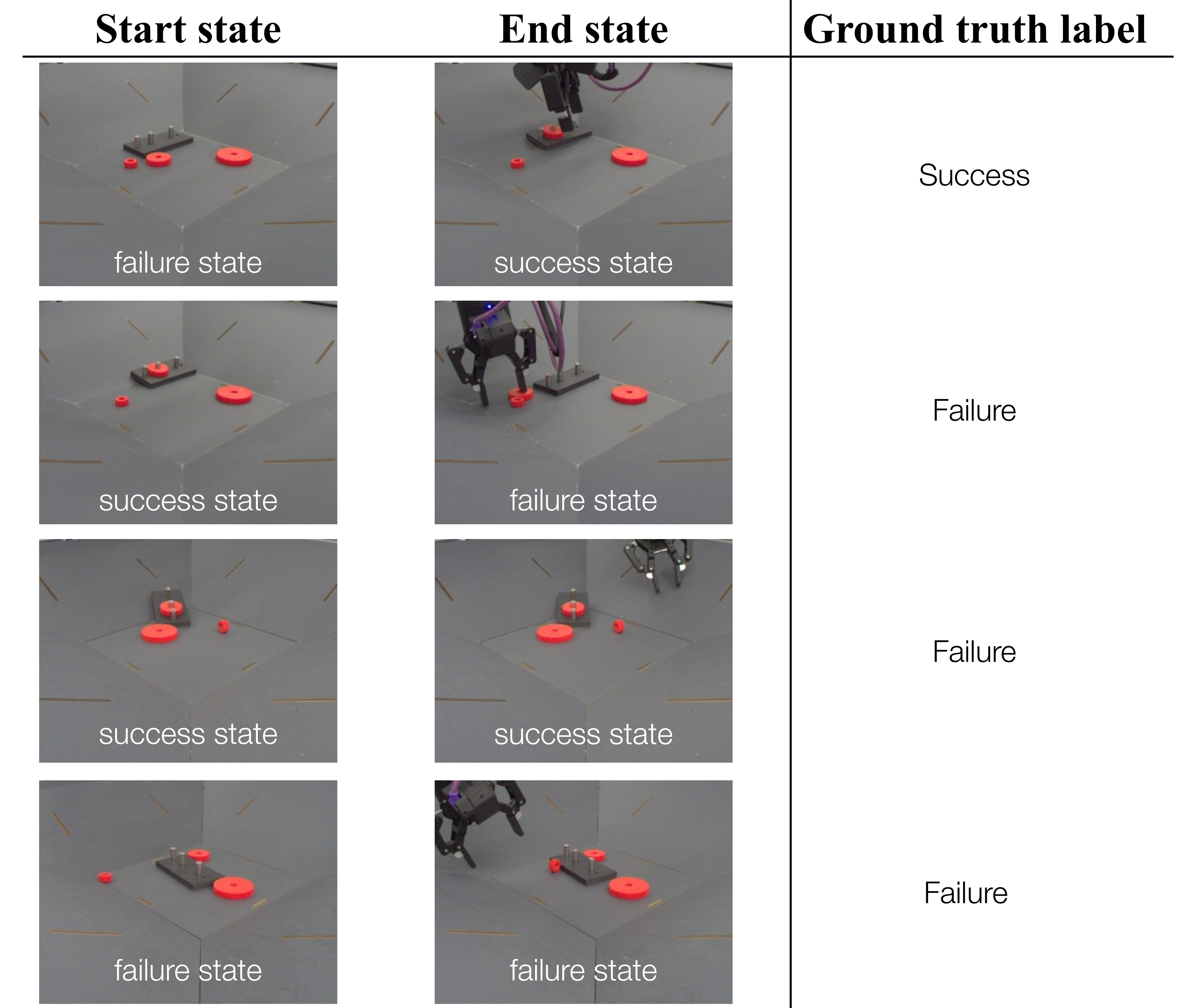}
    \caption{{\bf Ground truth labels for robotics tasks.} The episode is considered positive only when it starts in a failure state and ends in a success state, all other episodes are considered as negative.}
    \label{fig:GTrobotics}
\end{figure}

Figure \ref{fig:GTrobotics} shows how the ground truth success and failure labels are assigned to the full episodes. For an episode to be successful, it must start in a failure state and terminate in a success state.

\subsection{Data Efficiency in robotics domain}

We investigate whether the pretraining used for Flamingo makes it more amenable to accurate success detection in the lower-data regime. For this set of experiments, we train on only 100-200 episodes (100x less than the tens of thousands of episodes used in the above experiments) per task and evaluate on the same in-domain test set. As shown in Table \ref{tab:dataeff}, for five of the six tasks the Flamingo-based model is less affected by the smaller dataset than the ResNet-based model.

\begin{table}[ht!]
    \centering
    \footnotesize
    \begin{tabular}{c|c|c|c}
    \toprule
        Balanced Accuracy & Insert Small & Insert Medium & Insert Large \\ \midrule
        \baseline{} & 68.7\% \textcolor{red}{(-29.2\%)} & 70.2\% \textcolor{red}{(-28.3\%)} & 89.7\% \textcolor{red}{(-9.4\%)} \\ 
        \ours{}  & \textbf{77.6\%} \textcolor{red}{(-18.3)}& \textbf{85.3\%} \textcolor{red}{(-9.1\%)} & \textbf{93.2\%} \textcolor{red}{(-1.8\%)} \\ \midrule
         & Remove Small & Remove Medium & Remove Large \\ \midrule
         \baseline{}  &  \textbf{86.7\%} \textcolor{red}{(-10.6\%)} & \textbf{95.3\%} \textcolor{red}{(-3.4\%)} & \textbf{95.7\%} \textcolor{red}{(-2.7\%)} \\ 
       \ours{} & 70.5\% \textcolor{red}{(-11.6\%)} & 86.7\% \textcolor{green}{(+3.3\%)} & 87.1\% \textcolor{red}{(-0.0\%)} \\ \bottomrule
    \end{tabular}
    \caption{Data Efficiency -- train on 100-200 episodes, evaluate on 50-60k episodes. }
    \label{tab:dataeff}
\end{table}


\end{document}